\documentclass[runningheads]{llncs}
\usepackage{graphicx}
\usepackage{comment}
\usepackage{amsmath,amssymb} 
\usepackage{color}

\usepackage{url}
\usepackage[backref]{hyperref} 
\usepackage{graphicx}
\usepackage{amsmath}
\usepackage{booktabs}
\usepackage{algorithm}
\usepackage{algorithmic}
\usepackage{amssymb}
\usepackage{multirow}
\usepackage{graphicx}
\usepackage{colortbl}
\usepackage{tabu}
\usepackage{pifont}

\begin{document}
\pagestyle{headings}
\mainmatter
\def\ECCVSubNumber{5406}  

\title{Instance Adaptive Self-Training for Unsupervised Domain Adaptation} 

\titlerunning{Instance Adaptive Self-Training}
%
\author{Ke Mei\inst{1} \and
Chuang Zhu\inst{1} \thanks{The corresponding author: Chuang Zhu.} \and
Jiaqi Zou\inst{1} \and Shanghang Zhang\inst{2}}
%
\authorrunning{K. Mei et al.}
%
\institute{School of Information and Communication Engineering, Beijing University of Posts and Telecommunications, Beijing, China \\
\email{\{raykoo,czhu,jqzou\}@bupt.edu.cn} \\
\and
EECS, University of California, Berkeley, Berkeley, CA 94720, USA\\
\email{shz@eecs.berkeley.edu}}

\maketitle

\begin{abstract}
The divergence between labeled training data and unlabeled testing data is a significant challenge for recent deep learning models. Unsupervised domain adaptation (UDA) attempts to solve such a problem. Recent works show that self-training is a powerful approach to UDA. However, existing methods have difficulty in balancing scalability and performance. In this paper, we propose an instance adaptive self-training framework for UDA on the task of semantic segmentation. To effectively improve the quality of pseudo-labels, we develop a novel pseudo-label generation strategy with an instance adaptive selector. Besides, we propose the region-guided regularization to smooth the pseudo-label region and sharpen the non-pseudo-label region. Our method is so concise and efficient that it is easy to be generalized to other unsupervised domain adaptation methods. Experiments on `GTA5 to Cityscapes' and `SYNTHIA to Cityscapes' demonstrate the superior performance of our approach compared with the state-of-the-art methods.
\end{abstract}

\section{Introduction}
Domain shifts refer to the divergence between the training data (source domain) and the testing data (target domain), induced by factors such as the variance in illumination, object viewpoints, and image background \cite{tsai2018learning,chen2018domain}. Such domain shifts often lead to the phenomenon that the trained model suffers from a significant performance drop in the unlabeled target domain. The unsupervised domain adaptation (UDA) methods aim to improve the model generalization performance by transferring knowledge from labeled source domain to unlabeled target domain.

Recently, the adversarial training (AT) methods have received significant attention for semantic segmentation \cite{tsai2018learning,hoffman2018cycada,huang2018multimodal,long2018conditional,du2019ssf,vu2019advent}. These methods aim to minimize a series of adversarial losses to align source and target feature distributions. More recently, an alternative research line to reduce domain shift focuses on building schemes based on the self-training (ST) framework \cite{zou2018unsupervised,chen2019progressive,zheng2019unsupervised,zou2019confidence,lian2019constructing,zhang2019category}. These works iteratively train the model by using both the labeled source domain data and the generated pseudo-labels for the target domain and thus achieve the alignment between source and target domains. Besides, several works \cite{li2019bidirectional,tsai2019domain,zheng2019unsupervised} have explored to combine AT and ST methods, which shows great potential on semantic segmentation UDA. Through carefully designed network structure, these methods achieve state-of-the-art performance on the benchmark.

\begin{table}[ht]
\caption{Performance comparison of AT and ST. \emph{AT}: adversarial training based methods; \emph{ST}: self-training based methods; \emph{AT + ST}: the mixed methods}

\resizebox{1\linewidth}{!}{
\begin{tabular}{lcccccccc}
\toprule
Method   & BLF\cite{li2019bidirectional} & AdaptMR\cite{zheng2019unsupervised} & AdaptSeg\cite{tsai2018learning} & AdvEnt\cite{vu2019advent} & PyCDA\cite{lian2019constructing} & CRST\cite{zou2019confidence} & Ours & mean \\
\midrule
AT & 44.3 & 42.7 & 42.4 & 45.5 & - & - & 43.8 & 43.7 \\ 
ST & - & - & - & - & 47.4 & 47.1 & 48.8 & 47.8 \\ 
AT+ST & 48.5 & 48.3 & - & - & - & - & \textbf{50.2} & \textbf{49.0} \\
\bottomrule                  
\end{tabular}
}
\label{table:intro}
\end{table}

Despite the success of these AT and ST methods, a natural question comes up: what is the most effective one among these methods? AT or ST? Table \ref{table:intro} lists some of the above representative methods performance on the GTA5 to Cityscapes benchmark. All these methods use the same segmentation network for a fair comparison. In terms of performance, an explicit conclusion is: AT + ST (49.0) \cite{li2019bidirectional,zheng2019unsupervised} $>$ ST (47.8) \cite{lian2019constructing,zou2019confidence} $>$ AT (43.7) \cite{tsai2018learning,vu2019advent}. The mixed methods, such as BLF \cite{li2019bidirectional} and AdaptMR \cite{zheng2019unsupervised}, both have achieved great performance gains (+ 4.2, + 5.6) after using ST. However, in order to achieve better performance, these mixed methods generally have serious coupling between sub-modules (such as network structure dependency), thus losing scalability and flexibility.

This paper aims to propose a self-training framework for semantic segmentation UDA, which has good scalability that can be easily applied to other non-self-training methods and achieves state-of-the-art performance. To achieve this, we locate the main obstacle of existing self-training methods is how to generate high-quality pseudo-labels. This paper designs a new pseudo-label generation strategy and model regularization to solve this obstacle.

%
\begin{figure}[ht] 
    \centering 
    \includegraphics[width=3.8in]{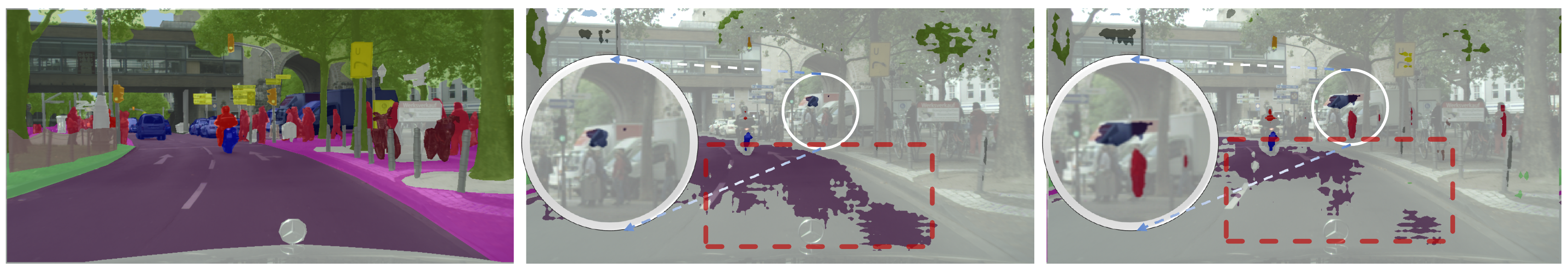} 
    \caption{Pseudo-label results. Columns correspond to original images with ground truth labels, class-balanced method, and our method}
    \label{fig:itro} 
\end{figure}

The pseudo-label generation suffers from information redundancy and noise. The generator tends to keep pixels with high confidence as pseudo-labels and ignore pixels with low confidence. Because of this conservative threshold selection, they are inefficient when more similar samples with high confidence are applied to training. The existing class-balanced self-training (CBST) \cite{zou2018unsupervised} utilized rank-based reference confidence for each class among all related images. This results in the ignorance of key information from the hard images with most of the pixels having low prediction scores. For example, in Fig. \ref{fig:itro}, the pseudo-labels generated by CBST are concentrated on the road, while pedestrians and trucks are ignored, which loses much learnable information. Therefore, we try to design a pseudo-label generation that can be adjusted adaptively according to the instance strategy to reduce data redundancy and increase the diversity of pseudo-labels.


%
\begin{figure}[ht] 
    \centering 
    \includegraphics[width=3.8in]{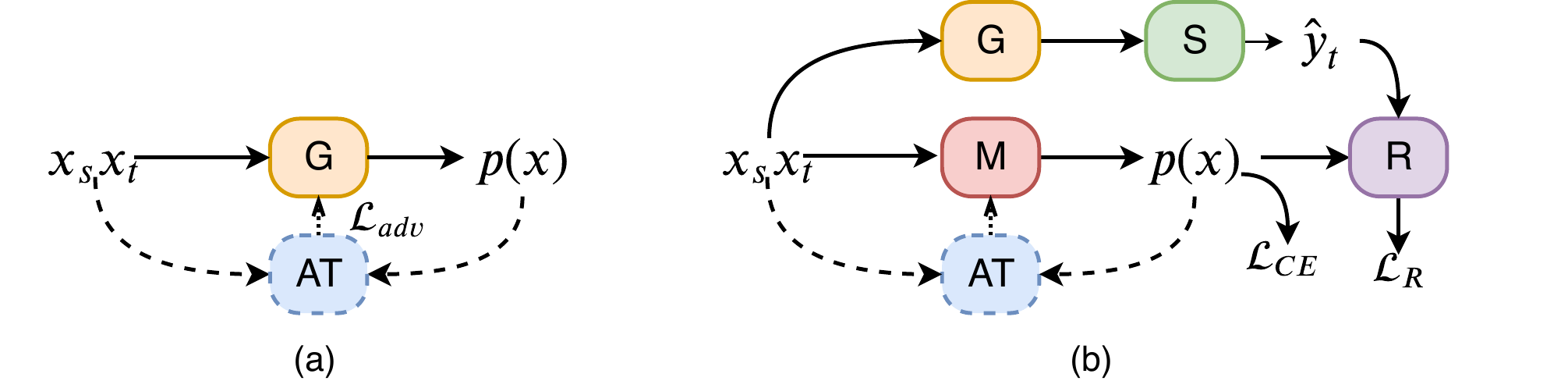} 
    \caption{IAST framework. (a) Warm-up phase, an initial model G is trained using any existing non-self-training method (eg. AT). (b) Self-training phase, the selector S filters the pseudo-labels generated by G, and R is the regularization }
    \label{fig:ust} 
\end{figure}

In this work, we propose an instance adaptive self-training framework (IAST) for semantic segmentation UDA, as shown in Fig. \ref{fig:ust}. We employ an instance adaptive selector in considering pseudo-label diversity during the training process. Besides, we design region-guided regularization in our framework, which has different roles in the pseudo-label region and the non-pseudo-label region. The main contributions of our work are summarized as follows:

\begin{itemize}
\item We propose a new self-training framework. Our methods significantly outperform the current state-of-the-art methods on the public semantic segmentation UDA benchmark.

\item We design an instance adaptive selector to involve more useful information for training. It effectively improves the quality of pseudo-labels. Besides, region-based regularization is designed to smooth the prediction of the pseudo-label region and sharpen the prediction of the non-pseudo-label region. 

\item We propose a general approach that makes it easy to apply other non-self-training methods to our framework. Moreover, our framework can also be extended to semi-supervised semantic segmentation tasks.
\end{itemize}

\section{Related Works}
\noindent\textbf{Adversarial training for UDA:} A large number of UDA schemes \cite{baktashmotlagh2013unsupervised,kan2015bi,lee2019sliced,li2018multi} are proposed to reduce the domain gap by building shared embedding space to both the source and target domain. Following the same idea, many adversarial training based UDA methods are proposed by adding a domain discriminator in recent years \cite{tsai2018learning,huang2018multimodal,long2018conditional,du2019ssf,vu2019advent,zhao2017self,zhao2018towards}. With adversarial training, the domain adversarial loss can be minimized to directly align features between two domains. Motivated by the recent image-to-image translation works, some works \cite{hoffman2018cycada,li2019bidirectional} regard the mapping from the source domain to the target domain as the image synthesis problem that reduce the domain discrepancy before training.

~\\

\noindent\textbf{Self-training:} Self-training schemes are commonly used in semi-supervised learning (SSL) areas \cite{li2005setred}.These works iteratively train the model by using both the labeled source domain data and the generated pseudo-labels in the target domain and thus achieve the alignment between the source and target domain \cite{triguero2015self}. However, these methods directly choose pseudo-labels with high prediction confidence, which will result in the model bias towards easy classes and thus ruin the transforming performance for the hard classes. To solve this problem, the authors in \cite{zou2018unsupervised} proposed a class-balanced self-training (CBST) scheme for semantic segmentation, which shows comparable domain adaptation performance to the best adversarial training based methods. \cite{lian2019constructing} proposed a self-motivated pyramid curriculum domain adaptation method using self-training. More recently, CRST \cite{zou2019confidence} further integrated a variety of confidence regularizers to CBST, producing better domain adaption results.

~\\

\noindent\textbf{Regularization:} Regularization refers to schemes that are intended to reduce the testing error and thus make the trained model generalize well to unseen data \cite{goodfellow2016deep,kukavcka2017regularization}. For deep neural network learning, different kinds of regularization schemes such as weight decay \cite{krizhevsky2012imagenet} and label smoothing \cite{szegedy2016rethinking} are proposed. The recent work \cite{zou2019confidence} designed labels and model regularization under self-training architecture for UDA. However, the proposed regularization scheme is just applied to the pseudo-label region.

\section{Preliminary}
\subsection{UDA for Semantic Segmentation}
It is assumed that there are two domains: source domain $S$ and target domain $T$. The source domain includes image $\mathbb{X}_{S}=\{x_{s}\}$, semantic mask $\mathbb{Y}_{S}=\{y_{s}\}$, and the target domain only has image $\mathbb{X}_{T}=\{x_{t}\}$. In UDA, the semantic segmentation model is trained only from the ground truth $\mathbb{Y}_{S}$ as the supervisory signal. UDA semantic segmentation model can be defined as follows:
$$\{\mathbb{X}_{S}, \mathbb{Y}_{S}, \mathbb{X}_{T}\}\Rightarrow \mathbf{M}_{UDA}$$ 
$\mathbf{M}_{UDA}$ uses some special losses and domain adaptation methods to align the distribution of two domains to learn domain-invariant feature representation.

\subsection{Self-training for UDA}
Because the ground truth labels of target domain are not available, we can treat the target domain as an extra unlabeled dataset. In this case, the UDA task can be transformed into a semi-supervised learning (SSL) task. Self-training is an effective method for SSL. The problem can be described as the following forms:
\begin{equation}
    \begin{aligned}
\min \limits_{\mathbf{w}} \mathcal{L}_{CE} =& - \frac{1}{\left | \mathbb{X}_{S} \right |} \sum_{\mathbf{x_{s}} \in \mathbb{X}_{S}} \sum_{c=1}^{C}y_{s}^{(c)}\log p(c|\mathbf{x}_{s}, \mathbf{w}) \\
& - \frac{1}{\left | \mathbb{X}_{T} \right |} \sum_{\mathbf{x}_{t} \in \mathbb{X}_{T}} \sum_{c=1}^{C}\hat{y}_{t}^{(c)}\log p(c|\mathbf{x}_{t}, \mathbf{w})
 \label{eq:1}
 \end{aligned}
\end{equation}
where $C$ is the number of classes, $y_{s}^{(c)}$ indicates the label of class $c$ in source domain, and $\hat{y}_{t}^{(c)}$ indicates the pseudo-label of class $c$ in target domain. $\mathbf{x}_{s}$ and $\mathbf{x}_{t}$ are input images, $\mathbf{w}$ indicates weights of $\mathbf{M}$,  $p(c|\mathbf{x},\mathbf{w})$ is the probability of class $c$ in  softmax output, and $\left | \mathbb{X} \right |$ indicates the number of images.

In particular, $\hat{\mathbb{Y}}_{T}=\{ \hat{y}_{t}\}$ are the ``pseudo-labels" generated according to the existing model, which is limited to a one-hot vector (only single 1 and all the others 0) or an all-zero vector.
The pseudo-labels can be used as approximate target ground truth labels.

Initially, pseudo-labels are generated before the training process. After this, Eq.\eqref{eq:1} can be used to directly minimize the cross-entropy loss of the source and target. Pseudo-labels are updated periodically during the self-training process.

\subsection{Adversarial training for UDA}
Adversarial training uses an additional discriminator to align feature distributions. The discriminator $\mathbf{D}$ attempts to distinguish the feature distribution in the output space of the source and target. The segmentation model $\mathbf{M}$ attempts to fool the discriminator to confuse the feature distributions of the source and target, thereby aligning the feature distributions. The optimization process is as follows:

\begin{equation}
    \begin{aligned}
\min \limits_{\mathbf{w}} \max \limits_{\mathbf{D}} \mathcal{L}X_{AT} = &- \frac{1}{\left | \mathbb{X}_{S} \right |} \sum_{\mathbf{x}_{s} \in \mathbb{X}_{S}} \sum_{c=1}^{C}y_{s}^{(c)}\log p(c|\mathbf{x}_{s}, \mathbf{w}) \\
&+ \frac{\lambda_{adv}}{\left | \mathbb{X}_{T} \right |} \sum_{\mathbf{x}_{t} \in \mathbb{X}_{T}} \left [ \mathbf{D}(\mathbf{M}(\mathbf{x}_{t}, \mathbf{w})) - \boldsymbol{1}  \right ]^{2}
\label{eq:2}
 \end{aligned}
\end{equation}

The first term is the cross-entropy loss of source, and the second term uses a mean squared error as the adversarial loss, where $\lambda_{adv}$ is the weight of the adversarial loss. Eq. \eqref{eq:2} is used to optimize $\mathbf{M}$ and $\mathbf{D}$ alternately.

\section{Proposed Method}

\begin{figure}[ht] 
    \centering 
    \includegraphics[width=4.0in]{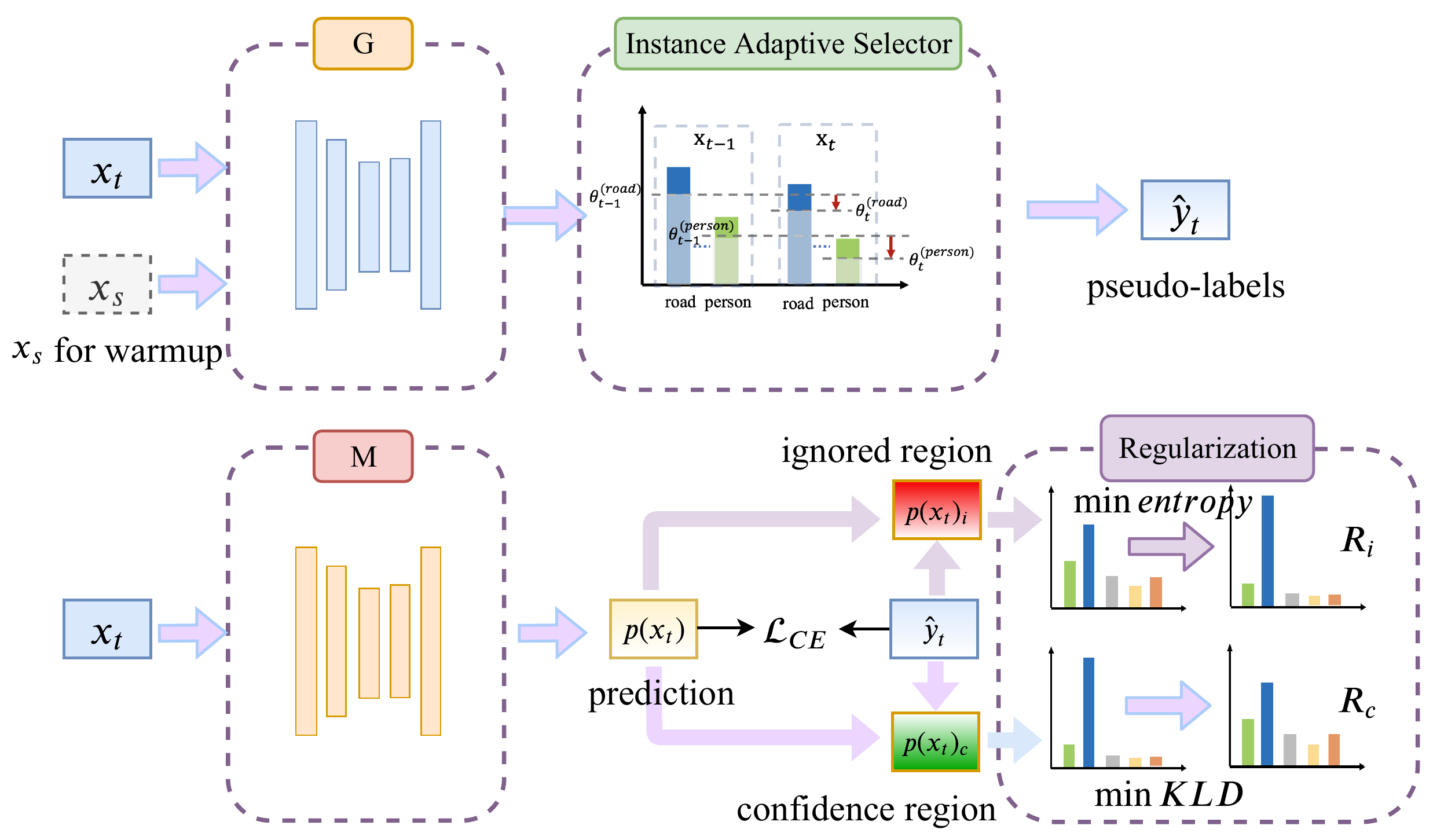} 
    \caption{Proposed IAST framework overview}
    \label{fig:framework} 
\end{figure}

An overview of our framework is shown in Fig. \ref{fig:framework}. We propose an instance adaptive self-training framework (IAST) with instance adaptive selector (IAS) and region-guided regularization. IAS selects an adaptive pseudo-label threshold for each semantic category in units of images and dynamically reduces the proportion of ``hard" classes, to eliminate noise in the pseudo-labels. Besides, region-guided regularization is designed to smooth the prediction of the confident region and sharpen the prediction of the ignored region. Our overall objective function is as follows:

\begin{equation} \label{eq:all}
    \begin{aligned}
        &\min \limits_{\mathbf{w}}  \mathcal{L}_{CE}(\mathbf{w},\hat{\mathbb{Y}}_{T}) + \mathcal{L}_{R}(\mathbf{w}) \\ 
        & =\mathcal{L}_{CE}(\mathbf{w},\hat{\mathbb{Y}}_{T}) 
        + (\lambda_{i}\mathcal{R}_{i}(\mathbf{w})  
        + \lambda_{c}\mathcal{R}_{c}(\mathbf{w}))
 \end{aligned}
\end{equation}
where $\mathcal{L}_{CE}$ is the cross-entropy loss, which is different from Eq.\eqref{eq:1} and only calculates the cross-entropy loss of the target domain images. $ \hat{\mathbb{Y}}_{T}$ is the set of pseudo-labels, and the detailed generation process is described in Section~\ref{subsection:p}. $\mathcal{R}_{i}$ and $\mathcal{R}_{c}$ are regularization of the ignored and confidence regions, which is described in Section~\ref{subsection: r}. And $\lambda_{i}$, $\lambda_{c}$ are regularization weights.

The IAST training process consists of three phases.
\begin{itemize}
    \item (a) In the \emph{warm-up phase}, a non-self-training method uses both the source data and the target data to train an initial segmentation model $\mathbf{M}_0$  as the initial pseudo-label generator $\mathbf{G}_0$.
    \item (b) In the \emph{pseudo-label generation phase}, $\mathbf{G}$ is used to obtain the prediction result of the target data, and a pseudo-label is generated by an instance adaptive selector.
    \item (c) In the \emph{self-training phase}, according to Eq.\eqref{eq:all}, the segmentation model $\mathbf{M}$ is trained using the target data.
\end{itemize}

\noindent\textbf{Why warm-up?} Before self-training, we expect to have a stable pre-trained model so that IAST can be trained in the right direction and avoid disturbances caused by constant fitting the noise of pseudo-labels. We use the adversarial training method described in Section 3.3 to obtain a stable model by roughly aligning the output of the source and target. In addition, in the warm-up phase, we can optionally apply any other semantic segmentation UDA method as the basic method, and it can be retained even in the (c) phase. In fact, we can use IAST as a decorator to decorate other basic methods.\\

\noindent\textbf{Multi-round self-training.} Performing (b) phase and (c) phase once counts as one round. As with other self-training tasks, in this experiment we performed a total of three rounds. At the end of each round, the parameters of model $\mathbf{M}$ will be copied into model $\mathbf{G}$ to generate better target domain prediction results in the next round.

\begin{figure}[ht] 
    \centering 
    \includegraphics[width=4.5in]{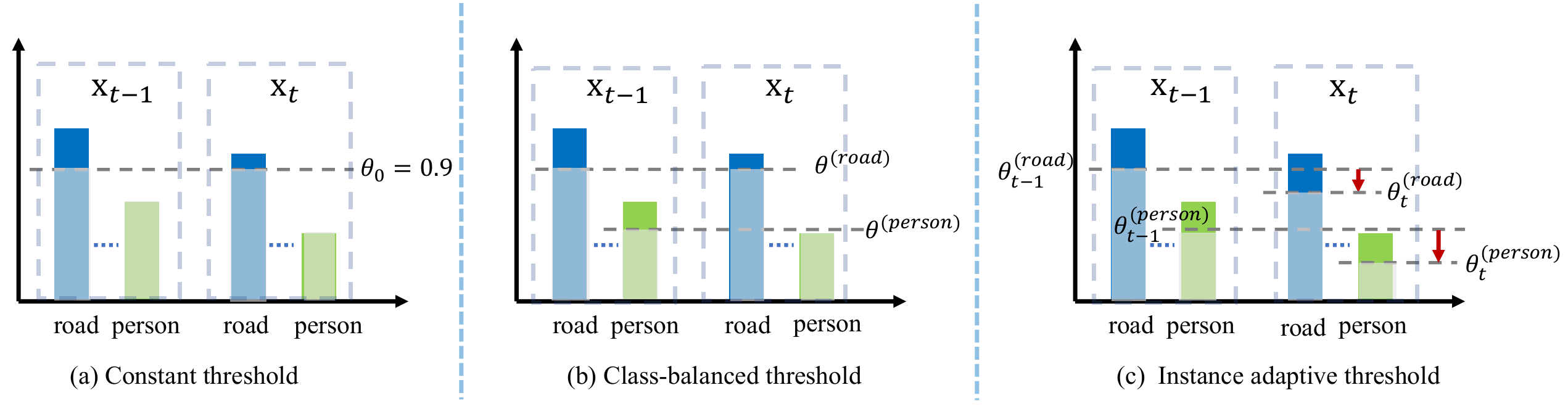} 
    \caption{Illustration of three different thresholding methods. $\mathbf{x}_{t-1}$ and $\mathbf{x}_{t}$ represent two consecutive instances, the bars approximately represent the probabilities of each class. (a) A constant threshold is used for all instances. (b)  class-balanced thresholds are used for all instances. (c) Our method adaptively adjusts the threshold of each class based on the instance}
    \label{fig:pseudo-methods} 
\end{figure}

\subsection{Pseudo-Label Generation Strategy with an Instance Adaptive Selector} \label{subsection:p}

 Pseudo-labels $\hat{\mathbb{Y}}_{T}$ have a decisive effect on the quality of self-training. The generic pseudo-label generation strategy can be simplified to the following form when segmentation model parameter $\mathbf{w}$ is fixed:

\begin{equation}
    \begin{aligned}
        \min \limits_{\hat{\mathbb{Y}}_{T}} & -  \frac{1}{\left | \mathbb{X}_{T} \right |} \sum_{\mathbf{x_{t}} \in \mathbb{X}_{T}} \sum_{c=1}^{C}\hat{y}_{t}^{(c)}
        \log\frac{p(c|\mathbf{x}_{t}, \mathbf{w})}
        {\theta_{t}^{(c)}}\\
        & s.t. \ \hat{\mathbf{y}}_{t} \in \{[onehot]^{C}\}\cup \mathbf{0} \ , \ \forall  \hat{\mathbf{y}}_{t} \in \hat{\mathbb{Y}}_{T}
 \end{aligned}
\end{equation}
where $\theta^{(c)}$ indicates the confidence threshold for class $c$, and $\mathbf{\hat{y}_{t}}=[\hat{y}_{t}^{(1)},...,\hat{y}_{t}^{(C)}]$ is required to be a one-hot vector or a all-zero vector. Therefore, $\hat{y}_{t}^{(c)}$ can be solved by Eq.\eqref{eq:y}.

\begin{equation}\label{eq:y}
    \hat{y}_{t}^{(c)}=\left\{
\begin{array}{rl}
1,       &   if \ c=\mathop{\arg\max} \limits_{c}p(c|\mathbf{x}_{t},\mathbf{w})\ and 
            \ p(c|\mathbf{x}_{t},\mathbf{w})> \theta^{(c)}\\
0,       & otherwise       
\end{array} \right.
\end{equation}

When class $c$ output probability $p(c|\mathbf{x}_{t},\mathbf{w})> \theta^{(c)}$, these pixels are regarded as confidence region (pseudo-label region), and the rest are ignored regions (non-pseudo-label region). Therefore, $\theta^{(c)}$ become the key to the pseudo-labels generation process. As shown in Fig.\ref{fig:pseudo-methods}: (a) the traditional pseudo-labels generation strategy based on a constant confidence threshold; (b) the generation strategy which uses the same class-balanced $\boldsymbol{\theta}$ for all target images; (c) we propose a data diversity-driven pseudo-labels generation strategy with an instant adaptive selector (IAS).

IAS maintains two thresholds $\{\boldsymbol{\theta}_t,\boldsymbol{\theta}_{\mathbf{x}_t} \}$, where $\boldsymbol{\theta}_t$ indicates the historical threshold and $\boldsymbol{\theta}_{\mathbf{x}_t}$ indicates the threshold of current instance $\mathbf{x}_t$. During the generation process, IAS dynamically updates  $\boldsymbol{\theta}_t$ based on  $\boldsymbol{\theta}_{\mathbf{x}_t}$ of the current instance $\mathbf{x}_t$, so each instance gets an adaptive threshold, combining global and local information. Specifically, for each instance $\mathbf{x}_t$, we  sort the confidence probability of each class in descending order, and then take the $\alpha \times 100\%$ confidence probability as the \textbf{local} threshold $\theta_{\mathbf{x}_{t}}^{(c)}$ for each class in instance $\mathbf{x}_t$. Finally, we use the exponentially weighted moving average to update the threshold $\boldsymbol{\theta}_{t}$ containing historical information as the \textbf{global} threshold. The details are summarized in Algorithm\ref{alg:algorithm1}.

\begin{algorithm}[tb]
\caption{pseudo-labels generation}
\label{alg:algorithm1}
\textbf{Input}: Model $\mathbf{M}$, target instance $\{\mathbf{x}_{t}\}^{T}$,  \\
\textbf{Parameter}: proportion $\alpha$, momentum $\beta$, weight decay $\gamma$,    \\
\textbf{Output}: target pseudo-labels 
\begin{algorithmic}[1] 
\STATE \textbf{init} $\boldsymbol{\theta}_{0}=\mathbf{0.9}$
\FOR{$t=1$ \TO $T$} 
\STATE $\mathbf{P}_{index}=\arg\max(\mathbf{M}(\mathbf{x}_{t}))$
\STATE $\mathbf{P}_{value}=\max(\mathbf{M}(\mathbf{x}_{t}))$
\FOR{$c=1$ \TO $C$}
\STATE $\mathbb{P}_{ \mathbf{x}_t}^{(c)}=\text{sort}(\mathbf{P}_{value}[\mathbf{P}_{index}=c],\text{descending})$
\STATE $\theta^{(c)}_{\mathbf{x}_{t}}=\Psi ( \mathbf{x}_{t},\theta _{t-1}^{(c)})$ $\ \ $ Eq.\eqref{eq:wd}
\ENDFOR
\STATE $\boldsymbol{\theta} _{t} = \beta \boldsymbol{\theta} _{t-1}+(1-\beta)\boldsymbol{\theta} _{\mathbf{x}_{t}}$ $\ \ $ Eq.\eqref{eq:ema}
\STATE $\hat{\mathbf{y}}_{t}=\text{onehot}(\mathbf{P}_{index}[\mathbf{P}_{value}>\boldsymbol{\theta} _{t}])$
\ENDFOR
\STATE \textbf{return} $\{\hat{\mathbf{y}}_{t}\}^T$
\end{algorithmic}
\end{algorithm}

\subsubsection{Exponential moving average (EMA) threshold.}When generating pseudo-labels one by one, we use an exponential moving average method, denoted as Eq.\eqref{eq:ema}, which can smooth the threshold of each instance, introduce past historical information, and avoid noise interference. Eq.\eqref{eq:wd} $\Psi (\mathbf{x}_{t},\theta _{t-1}^{(c)})$ represents the threshold for acquiring the current instance $\mathbf{x}_t$. $\beta$ is a momentum factor used to preserve past threshold information. As $\beta$ increases, the threshold $\theta _{t}^{(c)}$ becomes smoother.

\begin{equation}\label{eq:ema}
    \begin{aligned} 
        \theta _{t}^{(c)}  = \beta \theta _{t-1}^{(c)}+(1-\beta)\Psi ( \mathbf{x}_{t},\theta _{t-1}^{(c)})
    \end{aligned}
\end{equation}

\begin{equation}\label{eq:wd}
        \Psi (\mathbf{x}_{t},\theta _{t-1}^{(c)}) = \mathbb{P}_{ \mathbf{x}_t}^{(c)}\left [  \alpha  {\theta _{t-1}^{(c)}}^{\gamma}  | \mathbb{P}_{\mathbf{x}_t}^{(c)}  | \right ]
\end{equation}

\subsubsection{``Hard" classes weight decay (HWD).}For ``hard" classes, pseudo-labels tend to bring more noise labels. In Eq.\eqref{eq:wd}, we design $ {\theta_{t-1}^{(c)}}^{\gamma}$ to modify the proportion of pseudo-labels $\alpha$. $\gamma $ is a weight decay parameter, which is used to control the decay degree. The thresholds ${\theta_{t-1}^{(c)}} $ of the ``hard" classes are usually smaller, so HWD reduces more pseudo-labels of ``hard" classes. On the contrary the thresholds ${\theta _{t-1}^{(c)}}$ of easy classes is usually larger, so HWD has a weaker impact. It is easy to prove that when $\Psi (\mathbf{x}_{t},\theta _{t-1}^{(c)})=\theta _{t-1}^{(c)}$, $\theta$ will converge to a larger value, thereby reduce the amount of the ``hard" classes.



\subsection{Region-Guided Regularization} \label{subsection: r}
\subsubsection{Confident region KLD minimization.} During training, the model is prone to overfit pseudo-labels, which will damage the model. For the confidence region $\mathbb{I}_{\mathbf{x}_{t}}=\{\mathbf{1}\ | \ \mathbf{\hat{y}}_{t}^{(h,w)}>\mathbf{0} \}$, there are pseudo labels as supervising signals to supervise the model for learning. However, as shown in Table \ref{tab:gamma}, although a series of techniques for generating high-confidence pseudo labels have been used, the quality of the pseudo labels is still not as good as the ground truth labels, which means that there are some noise labels in the pseudo-labels. How to reduce the impact of noise labels is a key issue. Zou et al.  \cite{zou2019confidence} has proposed various regularization for this. We use the KLD which works best in \cite{zou2019confidence} to smooth the prediction results of the confidence region, so that the prediction results do not overfit the pseudo-labels.

\begin{equation} \label{eq:kld}
    \begin{aligned}
        \mathcal{R}_{c}= - \frac{1}{\left | \mathbb{X}_{T} \right |} \sum_{\mathbf{x}_{t} \in \mathbb{X}_{T}}\mathbb{I}_{\mathbf{x}_{t}} \sum_{c=1}^{C}\frac{1}{C}\log p(c|\mathbf{x}_{t}, \mathbf{w})
 \end{aligned}
\end{equation}

As shown in Eq.\eqref{eq:kld}, when the prediction result $\log p(c|\mathbf{x}_{t}, \mathbf{w})$ is approximately close to the uniform distribution (the probability of each class is $ \frac{1}{C} $), $\mathcal{R}_{c}$ gets smaller. KLD minimization promotes smoothing of confidence regionsand avoid the model blindly trusting false labels.

\subsubsection{Ignored region entropy minimization.}On the other hand, for the ignored region $\mathbb{I}_{\mathbf{x}_{t}}^{\complement}=\{\mathbf{1}\ | \ \mathbf{\hat{y}}_{t}^{(h,w)}=\mathbf{0} \}$, there is no supervision signal during the training process. Because the prediction result of the region $\mathbb{I}_{\mathbf{x}_{t}}^{\complement}$ is smooth and has low confidence, we use the minimized entropy of the ignored region to prompt the model to predict the low entropy result, which makes the prediction result look more ``sharper".

\begin{equation} \label{eq:ent}
    \begin{aligned}
        \mathcal{R}_{i}= - \frac{1}{\left | \mathbb{X}_{T} \right |} \sum_{\mathbf{x}_{t} \in \mathbb{X}_{T}}\mathbb{I}_{\mathbf{x}_{t}}^{\complement} \sum_{c=1}^{C}p(c|\mathbf{x}_{t}, \mathbf{w})\log p(c|\mathbf{x}_{t}, \mathbf{w})
 \end{aligned}
\end{equation}

As shown in Eq.\eqref{eq:ent}, sharpening the prediction result of the ignored region by minimizing $\mathcal{R}_{i}$ can promote the model to learn more useful features from the ignored region without any supervised signal, which has also been proved to be effective for UDA in the work \cite{vu2019advent}.

\section{Experiment}
\subsection{Experimental Settings}
\noindent\textbf{Network architecture and datasets.} We adapt Deeplab-v2 \cite{chen2017deeplab}, which is widely used in the semantic segmentation UDA problem, as our basic network architecture. ResNet-101\cite{he2016deep} is selected as the backbone network of the model. All experiments in this work are carried out under this network architecture. We evaluate our UDA methods for semantic segmentation on the popular synthetic-to-real adaptation scenarios: (a) GTA5 \cite{richter2016playing} to Cityscapes \cite{cordts2016cityscapes}, (b) SYNTHIA \cite{ros2016synthia} to Cityscapes. The GTA5 dataset has 24966 images that are rendered from the GTA5 game and $19$ classes with Cityscapes. SYNTHIA dataset includes $9400$ images and $16$ common classes with Cityscapes. Cityscapes is split into training set, validation set, and testing set. Following the standard protocols in \cite{tsai2018learning}, we use the training set which has $2975$ images as the target dataset and use the validation dataset to evaluate our models with mIoU.

\noindent\textbf{Implementation details.} In our experiments, we implement IAST using PyTorch on an NVIDIA Tesla V100. The training images are randomly cropped and resized to $1024\times512$, the aspect ratio of the crop window is $2.0$, and the window height is randomly selected from $[341\sim950]$ for GTA5 and $[341\sim640]$ for SYNTHIA. All weights of batch normalization layers were frozen. Deeplab-v2 is pre-trained on ImageNet. In IAST, we adopt Adam with learning rate $2.5\times10^{-5}$, batch size $6$ for $4$ epochs. The pseud-label parameters $\alpha$, $\beta$, $\gamma$ are set to $0.2$, $0.9$ and $8.0$. The regularization weights $\lambda_{i}$ and $\lambda_{c}$ are set to $3.0$ and $0.1$. Our code code and pre-trained molels are available at: \url{https://github.com/Raykoooo/IAST}


\begin{figure}[!htb] 
    \centering 
    \includegraphics[width=2.5in]{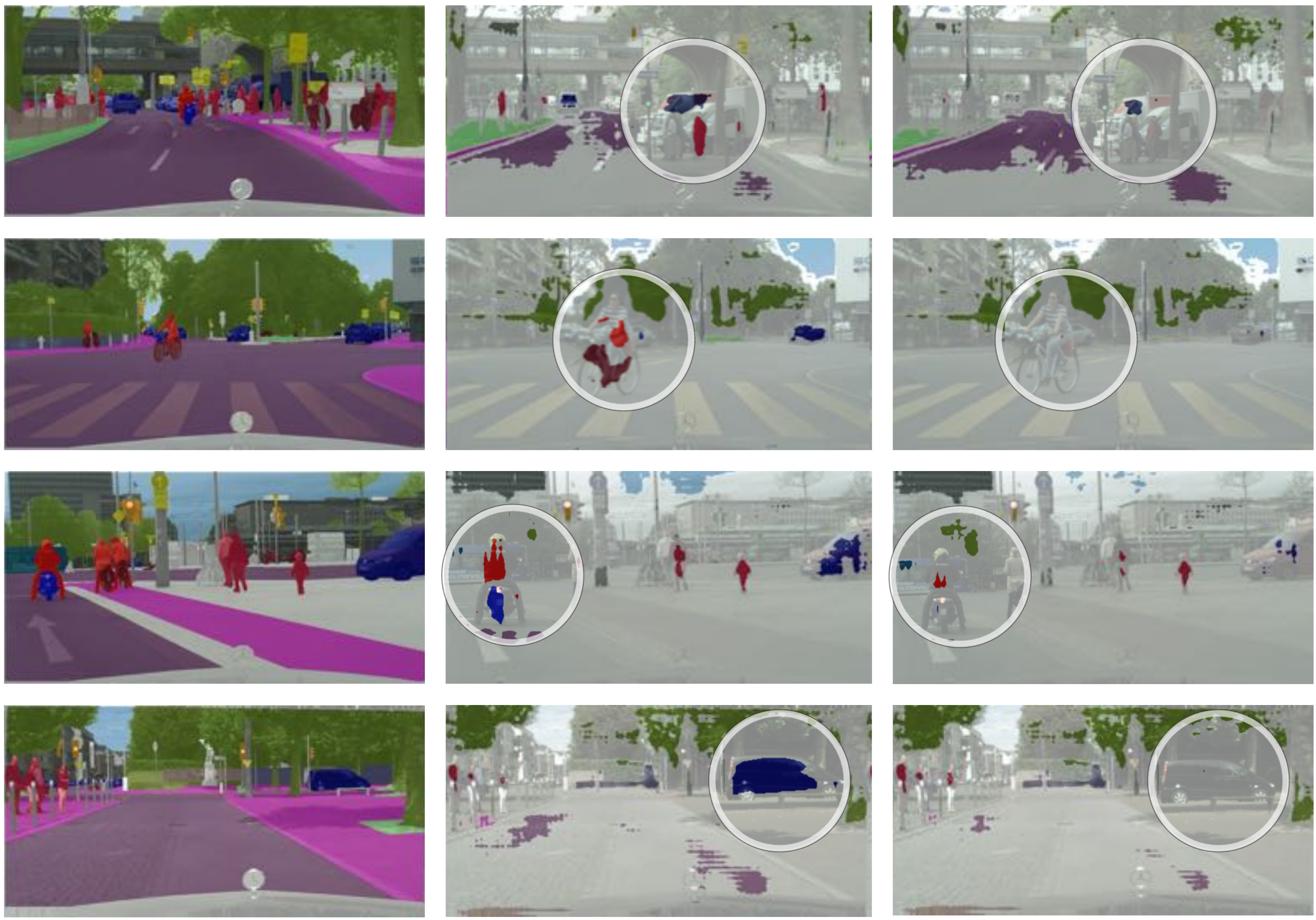} 
    \caption{Visualization of pseudo-labels. Columns correspond to original images with ground truth labels, our method, and class-balanced method \cite{zou2018unsupervised}}
    \label{fig:div} 
\end{figure}

\subsection{Discussion and Ablation Study}

\noindent\textbf{Why IAS works?} Table \ref{table:alpha} shows a sensitivity analysis on the parameter $\alpha$ and $\beta$. When we set $\alpha = 0.2$ and $\beta = 0$, it means IAS takes 20\% of each image as the confidence region. As a comparison, the class-balanced method \cite{zou2018unsupervised} takes 20\% of pixels in the whole target set as the confidence region. As shown in Fig. \ref{fig:div}, pseudo-labels of class-balanced method miss some pixels for persons, cars and bikes. In contrast, the pseudo-labels of our method are more diverse, especially for some ``hard" classes. When we set $\alpha = 0.2$ and $\beta = 0.9$, IAS combines global and local information to get more diverse content so that the model achieve the best performance.

\begin{table}[htb]
\scriptsize
\parbox{.46\linewidth}{
\caption{$\alpha$ and $\beta$ sensitivity analysis (GTA5 to Cityscapes)}
\resizebox{1\linewidth}{!}{
\begin{tabular}{cccc}
\toprule
\multicolumn{1}{c}{$\alpha$} & \multicolumn{1}{c}{$\beta$} & \multicolumn{1}{c}{Proportion(\%)} & \multicolumn{1}{c}{mIoU(\%)} \\ \midrule
.20                          & .0                            & 20.0                               & 49.8                         \\
.20                          & .50                         & 31.2                               & 50.3                         \\
.20                          & .90                          & 36.5                               & \textbf{50.5}                         \\
.20                         & .99                         & 40.1                               & 50.0                         \\\midrule
.30                          & .90                          & 42.5                               & 49.7                         \\
.50                          & .90                          & 48.6                               & 48.2                         \\ \midrule
\multicolumn{2}{l}{Constant(Fig. \ref{fig:pseudo-methods} a)}                                & 38.6                               & 45.1                         \\
\multicolumn{2}{l}{Class-balanced(Fig. \ref{fig:pseudo-methods} b)}                       & 20.0                               & 47.9           \\ \bottomrule
\end{tabular}
}
\label{table:alpha}
}
\hfill
\parbox{.50\linewidth}{
\scriptsize
\centering
\caption{$\lambda_i$ and $\lambda_c$ sensitivity analysis (GTA5 to Cityscapes)}
\setlength{\tabcolsep}{3mm}{
\begin{tabular}{ccc}
\toprule
\multicolumn{1}{c}{$\lambda_i$} & \multicolumn{1}{c}{$\lambda_c$} &  \multicolumn{1}{c}{mIoU(\%)} \\ \midrule
.5                          & .10                            & 50.6                         \\
1.0                          & .10                          & 51.1                         \\
2.0                          & .10                          & 50.9                         \\
3.0                          & .10                         & \textbf{51.5}                        \\
4.0                          & .10                          & 51.2                         \\
5.0                          & .10                          & 51.3                         \\ \midrule
3.0                          & .05                          & 50.6                         \\ 
3.0                          & .15                          & 51.0                         \\ 
\bottomrule
\end{tabular}
}
\label{table:lambda}
}
\end{table}

Fig.\ref{fig:dwd} shows that as the $\gamma$ increases, the proportion of some easy classes (sky, car) that have a high prediction score does not decrease significantly, while the proportion of some ``hard" classes (motor, wall, fence and pole) that have a low prediction score decreases sharply. This proves that Eq.\eqref{eq:wd} can effectively reduce the pseudo-labels of ``hard" classes and suppress noise interference in the pseudo-labels. Table \ref{tab:gamma} shows a sensitivity analysis on the parameter $\gamma$. We find that as the $\gamma$ increases, pseudo-labels have smaller proportions but have better quality. Therefore, we let $\gamma=8$ as the trade-off between the proportion and the quality of pseudo-labels. On the contrary, moderate regularization helps the model to improve the prediction accuracy and avoid overfitting the noise labels.

Table \ref{table:lambda} shows a sensitivity analysis of the parameter $\lambda_i$ and $\lambda_c$. We performed multiple sets of experiments with fixed $\lambda_i$ and $\lambda_c$, respectively. When $\lambda_c$ = 0.1 is fixed and $\lambda_i$ is gradually increased, the overall model performance tends to improve until $\lambda_i=4$. It can be expected that when the low entropy prediction is excessively performed in the non-pseudo-label region, the influence of noise will be amplified and the model will be damaged.


\makeatletter\def\@captype{figure}\makeatother
\begin{minipage}{.50\textwidth}
\includegraphics[width=2.2in]{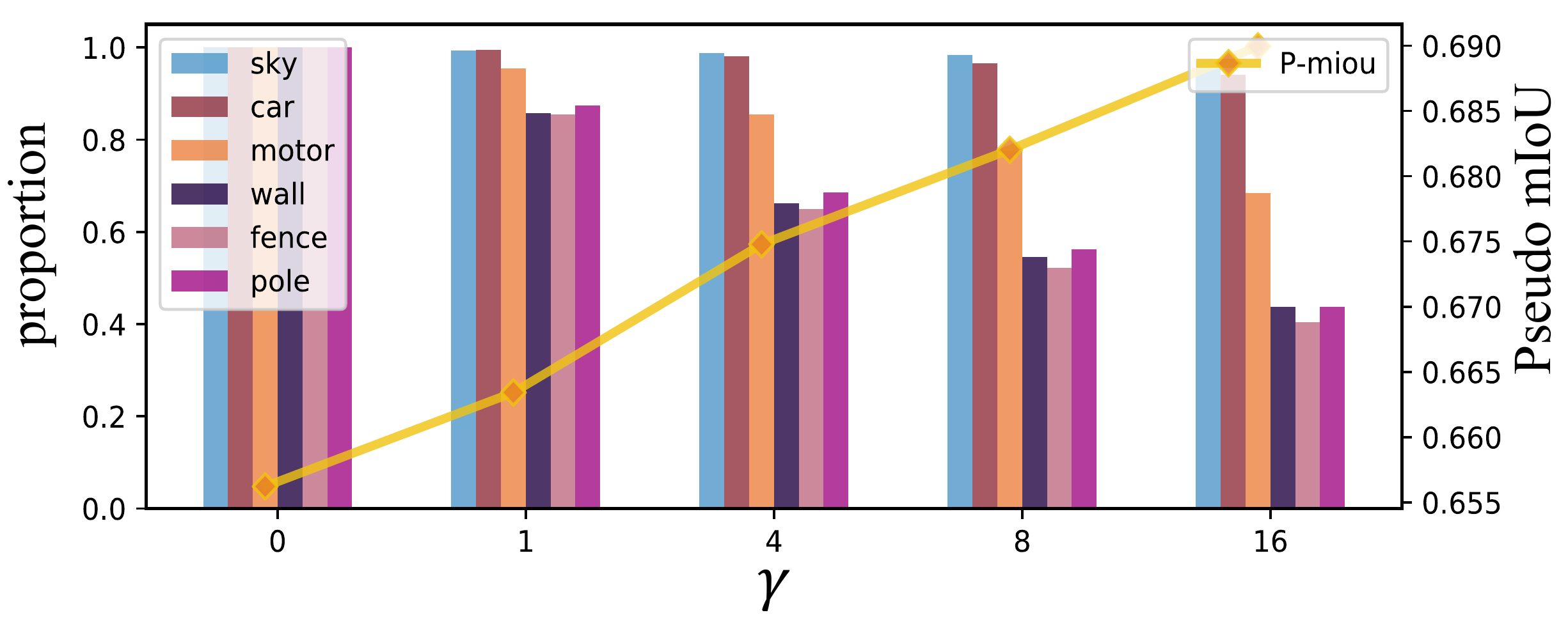}
\caption{Relationship between the pseudo-labels proportion and $\gamma$}
\label{fig:dwd} 
\end{minipage}
\makeatletter\def\@captype{table}\makeatother
\begin{minipage}{.38\textwidth}
\centering
\caption{$\gamma$ sensitivity analysis ($\alpha=0.2, \beta=1.0$). \emph{P-mIoU} means mIoU of pseudo-labels}
\resizebox{0.95\linewidth}{!}{
\setlength{\tabcolsep}{1.5mm}{
\begin{tabular}{cccccc}  
\toprule
$\gamma$  & 0 & 1 & 4 & 8 & 16 \\
\midrule
Proportion & 0.36 & 0.34 & 0.30 & 0.28 & 0.25 \\
P-mIoU(\%) & 65.6 & 66.3 & 67.4 & 68.2 & 69.0 \\
mIoU  (\%) & 50.5 & 50.8 & 51.2 & $\textbf{51.5}$ & 50.9 \\
\bottomrule
\end{tabular}
}
}
\label{tab:gamma}
\end{minipage}

~\\
\noindent\textbf{Ablation studies.} The results of the ablation studies are reported in Table \ref{table:ablation}. We attempt the methods proposed in Section~\ref{subsection:p} and Section~\ref{subsection: r} one by one to study their performance in the test set. From the data in Table \ref{table:ablation}, after using self-training (Fig. \ref{fig:pseudo-methods} a) without using any other techniques, the model performance has a gain of $1.3\%$. After adding IAST modules (IAS, $\mathcal{R}_i$, $\mathcal{R}_c$), the performance of the model is gradually and steadily improved, and finally, $51.5\%$ mIoU is achieved. In addition, we also try multi-scale testing and the combined result achieved the best $52.2\%$ mIoU.

\begin{table}[htb!]
\centering
\caption{Results of ablation study (GTA5 to Cityscapes)}
\resizebox{0.65\linewidth}{!}{
\setlength{\tabcolsep}{2.0mm}{
\begin{tabular}{lcccccrr}  
\toprule
Method & ST & IAS & $\mathcal{R}_c$ & $\mathcal{R}_i$ & mIoU & $\Delta$ \\
\midrule
Source & - &  - & - & -& 35.6 &0\\
Warm-up &  - & - & - & - & 43.8 & +8.2 \\
\midrule
+ Constant ST(Fig. \ref{fig:pseudo-methods} a) & \ding{51} &  &  &  &  45.1 & +1.3 \\
+ Instance adaptive selector  & \ding{51}& \ding{51} &   &  &  49.8 & +4.7\\
+ Confidence region R.  & \ding{51} &\ding{51} & \ding{51} &  &  50.7 & +0.9\\
+ Ignored region R.  & \ding{51} & \ding{51} &\ding{51}  & \ding{51} &  51.5 & +0.8\\
\bottomrule
\end{tabular}
}
}
\label{table:ablation}
\end{table}

\begin{table}[htb]
\caption{Results of our proposed method IAST and other state-of-the-art methods (GTA5 to Cityscapes). A\&S means a mixed method of AT and ST}
\resizebox{1\linewidth}{!}{
\renewcommand\arraystretch{1.3}
\begin{tabu}{l|c|rrrrrrrrrrrrrrrrrrr|r}
\tabucline[1.0pt]{-}
Method      & \rotatebox{90}{Arch.}                   &  \rotatebox{90}{Road} & \rotatebox{90}{SW}   & \rotatebox{90}{Build} & \rotatebox{90}{Wall} & \rotatebox{90}{Fence} & \rotatebox{90}{Pole}  & \rotatebox{90}{TL}    &  \rotatebox{90}{TS}   & \rotatebox{90}{Veg.} & \rotatebox{90}{Terrain} & \rotatebox{90}{Sky}  & \rotatebox{90}{PR}   & \rotatebox{90}{Rider} & \rotatebox{90}{Car}  & \rotatebox{90}{Truck} & \rotatebox{90}{Bus}  & \rotatebox{90}{Train} & \rotatebox{90}{Motor} & \rotatebox{90}{Bike} & mIoU \\ \hline
Source  \cite{tsai2018learning}    & \multirow{6}{*}{AT} & 75.8 & 16.8 & 77.2  & 12.5 & 21.0  & 25.5 & 30.1 & 20.1 & 81.3 & 24.6    & 70.3 & 53.8 & 26.4  & 49.9 & 17.2  & 25.9 & 6.5   & 25.3  & 36.0 & 36.6 \\
AdaptSegNet \cite{tsai2018learning} &                            & 86.5 & 36.0 & 79.9  & 23.4 & 23.3  & 23.9 & 35.2 & 14.8 & 83.4 & 33.3    & 75.6 & 58.5 & 27.6  & 73.7 & 32.5  & 35.4 & 3.9   & 30.1  & 28.1 & 42.4 \\
SIBAN   \cite{luosignificance}    &                            & 88.5 & 35.4 & 79.5  & 26.3 & 24.3  & 28.5 & 32.5 & 18.3 & 81.2 & 40.0    & 76.5 & 58.1 & 25.8  & 82.6 & 30.3  & 34.3 & 3.4   & 21.6  & 21.5 & 42.6 \\
SSF-DAN \cite{du2019ssf}     &                            & 90.3 & 38.9 & 81.7  & 24.8 & 22.9  & 30.5 & 37.0 & 21.2 & 84.8 & 38.8    & 76.9 & 58.8 & 30.7  & 85.7 & 30.6  & 38.1 & 5.9   & 28.3  & 36.9 & 45.4 \\
AdvEnt   \cite{vu2019advent}   &                            & 89.4 & 33.1 & 81.0  & 26.6 & 26.8  & 27.2 & 33.5 & 24.7 & 83.9 & 36.7    & 78.8 & 58.7 & 30.5  & 84.8 & 38.5  & 44.5 & 1.7   & 31.6  & 32.4 & 45.4 \\ 
APODA    \cite{yang2019adversarial}   &                            & 85.6 & 32.8 & 79.0  & 29.5 & 25.5  & 26.8 & 34.6 & 19.9 & 83.7 & 40.6    & 77.9 & 59.2 & 28.3  & 84.6 & 34.6  & 49.2 & 8.0   & 32.6  & 39.6 & 45.9 \\\hline
Source   \cite{zou2018unsupervised}   & \multirow{4}{*}{ST}          & 71.3 & 19.2 & 69.1  & 18.4 & 10.0  & 35.7 & 27.3 & 6.8  & 79.6 & 24.8    & 72.1 & 57.6 & 19.5  & 55.5 & 15.5  & 15.1 & 11.7  & 21.1  & 12.0 & 33.8 \\ 
CBST     \cite{zou2018unsupervised}   &                            & 91.8 & 53.5 & 80.5  & 32.7 & 21.0  & 34.0 & 28.9 & 20.4 & 83.9 & 34.2    & 80.9 & 53.1 & 24.0  & 82.7 & 30.3  & 35.9 & 16.0  & 25.9  & 42.8 & 45.9 \\
PyCDA\cite{lian2019constructing}   &                            & 90.5 & 36.3 & 84.4  & 32.4 & 28.7  & 34.6 & 36.4 & 31.5 & \textbf{86.8} & 37.9    & 78.5 & 62.3 & 21.5  & 85.6 & 27.9  & 34.8 & 18.0  & 22.9  & \textbf{49.3} & 47.4 \\
MRKLD     \cite{zou2019confidence}  &                            & 91.0 & 55.4 & 80.0  & 33.7 & 21.4  & \textbf{37.3} & 32.9 & 24.5 & 85.0 & 34.1    & 80.8 & 57.7 & 24.6  & 84.1 & 27.8  & 30.1 & \textbf{26.9}  & 26.0  & 42.3 & 47.1 \\ \hline
BLF   \cite{li2019bidirectional}      & \multirow{3}{*}{A\&S } & 91.0 & 44.7 & 84.2  & 34.6 & 27.6  & 30.2 & 36.0 & 36.0 & 85.0 & \textbf{43.6}    & 83.0 & 58.6 & 31.6  & 83.3 & 35.3  & 49.7 & 3.3   & 28.8  & 35.6 & 48.5 \\
AdaptMR  \cite{zheng2019unsupervised}   &                            & 90.5 & 35.0 & 84.6  & 34.3 & 24.0  & 36.8 & \textbf{44.1} & \textbf{42.7} & 84.5 & 33.6    & 82.5 & \textbf{63.1} & \textbf{34.4}  & 85.8 & 32.9  & 38.2 & 2.0   & 27.1  & 41.8 & 48.3 \\  
PatchAlign \cite{tsai2019domain} &                            & 92.3 & 51.9 & 82.1  & 29.2 & 25.1  & 24.5 & 33.8 & 33.0 & 82.4 & 32.8    & 82.2 & 58.6 & 27.2  & 84.3 & 33.4  & 26.3 & 2.2   & 29.5  & 32.3 & 46.5 \\ \hline
Source(ours)      & \multirow{3}{*}{A\&S} & 64.8 & 21.7 & 74.3  & 15.4 & 21.2  & 18.2 & 30.7 & 13.0 & 80.9 & 33.7    & 76.3 & 55.6 & 20.0  & 43.9 & 27.0  & 35.5 & 4.4   & 24.9  & 14.3 & 35.6 \\ 
\textbf{IAST(ours)}      &                            & 93.8 & 57.8 & 85.1  & 39.5 & 26.7  & 26.2 & 43.1 & 34.7 & 84.9 & 32.9    & 88.0 & 62.6 & 29.0  & 87.3 & 39.2  & 49.6 & 23.2  & 34.7  & 39.6 & 51.5 \\
\textbf{IAST-MST(ours)}  &                            & \textbf{94.1} & \textbf{58.8} & \textbf{85.4}  & \textbf{39.7} & \textbf{29.2}  & 25.1 & 43.1 & 34.2 & 84.8 & 34.6    & \textbf{88.7} & 62.7 & 30.3  & \textbf{87.6} & \textbf{42.3}  & \textbf{50.3} & 24.7  & \textbf{35.2}  & 40.2 & \textbf{52.2} \\ \tabucline[1.0pt]{-}
\end{tabu}
}
\label{table:gta5}
\end{table}

\begin{table}[!htb]
\centering
\caption{Results of our proposed method IAST and other state-of-the-art methods (SYNTHIA to Cityscapes)}
\resizebox{1.0\linewidth}{!}{
\renewcommand\arraystretch{1.3}
\begin{tabu}{l|c|rrrrrrrrrrrrrrrr|r|r}
\tabucline[1.0pt]{-}
Method      & \rotatebox{90}{Arch.}  &               \rotatebox{90}{Road} & \rotatebox{90}{SW}   & \rotatebox{90}{Build} & \rotatebox{90}{Wall*} & \rotatebox{90}{Fence*} & \rotatebox{90}{Pole*}  & \rotatebox{90}{TL}    &  \rotatebox{90}{TS}   & \rotatebox{90}{Veg.} & \rotatebox{90}{Sky}  & \rotatebox{90}{PR}   & \rotatebox{90}{Rider} & \rotatebox{90}{Car}  & \rotatebox{90}{Bus}  & \rotatebox{90}{Motor} & \rotatebox{90}{Bike} & mIoU & mIoU* \\ \hline
Source  \cite{tsai2018learning}    & \multirow{6}{*}{AT} & 55.6 & 23.8 & 74.6  & -     & -      & -     & 6.1  & 12.1 & 74.8 & 79.0 & 55.3 & 19.1  & 39.6 & 23.3 & 13.7  & 25.0 & -    & 38.6  \\  
AdaptSegNet \cite{tsai2018learning} &                            & 84.3 & 42.7 & 77.5  & -     & -      & -     & 4.7  & 7.0  & 77.9 & 82.5 & 54.3 & 21.0  & 72.3 & 32.2 & 18.9  & 32.3 & -    & 46.7  \\ 
SIBAN    \cite{luosignificance}    &                            & 82.5 & 24.0 & 79.4  & -     & -      & -     & 16.5 & 12.7 & 79.2 & 82.8 & 58.3 & 18.0  & 79.3 & 25.3 & 17.6  & 25.9 & -    & 46.3  \\  
SSF-DAN    \cite{du2019ssf}  &                            & 84.6 & 41.7 & 80.8  & -     & -      & -     & 11.5 & 14.7 & 80.8 & \textbf{85.3} & 57.5 & 21.6  & 82.0 & 36.0 & 19.3  & 34.5 & -    & 50.0  \\ 
AdvEnt   \cite{vu2019advent}      &                            & 85.6 & 42.2 & 79.7  & 8.7   & 0.4    & 25.9  & 5.4  & 8.1  & 80.4 & 84.1 & 57.9 & 23.8  & 73.3 & 36.4 & 14.2  & 33.0 & 41.2 & 48.0  \\ 
APODA     \cite{yang2019adversarial}     &                            & \textbf{86.4} & 41.3 & 79.3  & -     & -      & -     & 22.6 & 17.3 & 80.3 & 81.6 & 56.9 & 21.0  & 84.1 & \textbf{49.1} & 24.6  & 45.7 & -    & 53.1  \\ \hline
Source   \cite{zou2018unsupervised}   & \multirow{4}{*}{ST} & 64.3 & 21.3 & 73.1  & 2.4   & 1.1    & 31.4  & 7.0  & 27.7 & 63.1 & 67.6 & 42.2 & 19.9  & 73.1 & 15.3 & 10.5  & 38.9 & 34.9 & 40.3  \\ 
CBST       \cite{zou2018unsupervised}  &                            & 68.0 & 29.9 & 76.3  & 10.8  & 1.4    & 33.9  & 22.8 & 29.5 & 77.6 & 78.3 & 60.6 & 28.3  & 81.6 & 23.5 & 18.8  & 39.8 & 42.6 & 48.9  \\ 
PyCDA       \cite{lian2019constructing}  &                            & 75.5 & 30.9& 83.3  & \textbf{20.8}  & 0.7    & 32.7  & 27.3 & 33.5 & \textbf{84.7} & 85.0 & 64.1 & 25.4  & 85.0 & 45.2 & 21.2  & 32.0 & 46.7 & 53.3  \\ 
MRKLD   \cite{zou2019confidence}     &                            & 67.7 & 32.2 & 73.9  & 10.7  & 1.6    & \textbf{37.4}  & 22.2 & \textbf{31.2} & 80.8 & 80.5 & 60.8 & 29.1  & 82.8 & 25.0 & 19.4  & 45.3 & 43.8 & 50.1  \\ \hline
BLF     \cite{li2019bidirectional}    & \multirow{3}{*}{A\&S} & 86.0 & \textbf{46.7} & 80.3  & -     & -      & -     & 14.1 & 11.6 & 79.2 & 81.3 & 54.1 & 27.9  & 73.7 & 42.2 & 25.7  & 45.3 & -    & 51.4  \\
AdaptMR   \cite{zheng2019unsupervised}  &                            & 83.1 & 38.2 & 81.7  & 9.3   & 1.0    & 35.1  & 30.3 & 19.9 & 82.0 & 80.1 & 62.8 & 21.1  & 84.4 & 37.8 & 24.5  & \textbf{53.3} & 46.5 & 53.8  \\ 
PatchAlign  \cite{tsai2019domain} &                            & 82.4 & 38.0 & 78.6  & 8.7   & 0.6    & 26.0  & 3.9  & 11.1 & 75.5 & 84.6 & 53.5 & 21.6  & 71.4 & 32.6 & 19.3  & 31.7 & 40.0 & 46.5  \\ \hline
Source(ours)      & \multirow{2}{*}{A\&S} & 63.4 & 24.1 & 66.7  & 7.1   & 0.1    & 28.4  & 11.6 & 16.8 & 77.0 & 74.6 & 60.4 & 20.5  & 75.6 & 22.0 & 14.4  & 21.2 & 36.5 & 42.2  \\ 
\textbf{IAST(ours)}      &                            & 81.9 & 41.5 & \textbf{83.3}  & 17.7  & \textbf{4.6}    & 32.3  & \textbf{30.9} & 28.8 &  83.4 & 85.0 & \textbf{65.5} & \textbf{30.8}  & \textbf{86.5} & 38.2 & \textbf{33.1}  & 52.7 & \textbf{49.8} & \textbf{57.0}  \\
\tabucline[1.0pt]{-}
\end{tabu}
}
\label{table:synthia}
\end{table}

\subsection{Experimental Results}
\noindent\textbf{Comparison with the state-of-the-art methods:}The results of IAST and some other state-of-the-art methods on GTA5 to Cityscapes are present in Table\ref{table:gta5}. From the overall results, IAST has the best mIoU $52.2\%$ and has obvious advantages over other methods. Compared with some adversarial training methods AdaptSegNet \cite{tsai2018learning} and SIBAN \cite{luosignificance}, IAST improves by $9.6\%$ mIoU and have significant gains in almost all classes. Compared with the same self-training methods such as MRKLD \cite{zou2019confidence}, IAST improves by $4.8\%$ mIoU. In addition, BLF \cite{li2019bidirectional} is a method that combines adversarial training and self-training, which has the second-best $48.5\%$ mIoU. Compared to BLF, IAST still has a significant improvement.

Table \ref{table:synthia} is the results of the SYNTHIA to Cityscapes dataset. For a comprehensive comparison, as in the previous work, we also report two mIoU metrics: $13$ classes of mIoU* and $16$ classes of mIoU. The domain gap between SYNTHIA and Cityscapes is much larger than the domain gap between GTA5 and Cityscapes. Many of the methods that performed well on GTA5 to Cityscapes have experienced a significant performance degradation on this dataset. Correspondingly, the performance gap between different methods is becoming more apparent. IAST also achieves the best results, which are $49.8\%$ mIoU and $57.0\%$ mIoU* and significantly higher than all recent state-of-the-art methods.

\begin{table}
\small
\parbox{.40\linewidth}{
\centering
\caption{Semi-supervised learning results on the Cityscapes val set. 1/8, 1/4 and 1/2 mean the proportion of labeled images}

\resizebox{1\linewidth}{!}{
\begin{tabular}{lcccc}
\toprule
\multirow{2}{*}{Method} & \multicolumn{4}{c}{Data Amount}                                                                                                    \\ \cmidrule{2-5} 
                        & \multicolumn{1}{l}{1/8} & \multicolumn{1}{l}{1/4} & \multicolumn{1}{l}{1/2} & \multicolumn{1}{l}{Full} \\ \midrule
Baseline                & 57.3                    & 59.0                    & 61.2                                        &         70.2             \\
Univ-full\cite{kalluri2019universal}       & 55.9                    & -                       & -                                               & -                        \\
AdvSemi\cite{hung2019adversarial}        & 58.8                    & 62.3                    & 65.7                                         & 67.7                     \\
IAST(ours)                    & \textbf{64.6}                    & \textbf{66.7}                    & \textbf{69.8}                                   &  \textbf{70.2}                     \\ \bottomrule
\end{tabular}
}
\label{table:semi}
}
\hfill
\parbox{.55\linewidth}{
\centering
\caption{Extension analysis, applying IAST to non-self-learning UDA methods \cite{tsai2018learning,vu2019advent} (test on Cityscapes), and \emph{Source} means training IAST without warmup}

\resizebox{1\linewidth}{!}{
\begin{tabular}{lccccccc}
\toprule
\multirow{2}{*}{Method} & \multicolumn{3}{c}{GTA5}                                                    &  & \multicolumn{3}{c}{SYNTHIA}                                   \\ \cmidrule{2-4} \cmidrule{6-8} 
                        & \multicolumn{1}{c}{Base} & \multicolumn{1}{c}{+IAST} & \multicolumn{1}{c}{$\Delta$} &  & \multicolumn{1}{c}{Base} & \multicolumn{1}{c}{+IAST} & \multicolumn{1}{c}{$\Delta$} \\ \midrule
AdaptSeg\cite{tsai2018learning}                & 42.4                       & 50.2                     & +7.8                      &  & 46.7                       & 54.7 & +8.0                      \\
AdvEnt\cite{vu2019advent}                  & 45.4                       & 49.8                     & +4.4                      &  & 48.0                       & 55.1 & +7.1                      \\
Source                  & 35.6                       & 48.8                     & +13.2                     &  & 42.2                       & 54.2 & +12.0                     \\ \bottomrule
\end{tabular}
}
\label{table:apply}
}
\end{table}

\noindent\textbf{Apply to other UDA methods.} Because IAST has no special structure or model dependencies, it can be directly used to decorate other UDA methods. We chose two typical adversarial training methods, AdaptSeg\cite{tsai2018learning} and AdvEnt\cite{vu2019advent} for experiments. As shown in Table \ref{table:apply}, these two methods have significantly improved performance under the IAST framework.

\noindent\textbf{Extension: other tasks.} The self-training method can also be applied to semi-supervised semantic segmentation task. We use the same configuration as \cite{hung2019adversarial} in Cityscapes for semi-supervised training with different proportions of data as labeled data. As shown in Table \ref{table:semi}, we have significantly better performance than \cite{hung2019adversarial} and \cite{kalluri2019universal}.

\section{Conclusions}

In this paper, we propose an instance adaptive self-training framework for semantic segmentation UDA. Compared with other popular UDA methods, IAST still has a significant improvement in performance. Moreover, IAST is a method with no model or special structure dependency, which means that it can be easily applied to other UDA methods with almost no additional cost to improve performance. In addition, IAST can also be applied to semi-supervised semantic segmentation tasks, which also achieves state-of-the-art performance. We hope this work will prompt people to rethink the potential of self-training on UDA or semi-supervised learning tasks.

\section*{Acknowledgement}
This work was supported in part by the Natural Science Foundation of Beijing Municipality under Grant 4182044.


\clearpage
%
%

\bibliographystyle{splncs04}
\bibliography{egbib}
\end{document}